\documentclass[10pt,journal]{IEEEtran}  

\IEEEoverridecommandlockouts                              

\usepackage{graphics} 
\usepackage{epsfig} 
\usepackage{mathptmx} 
\usepackage{times} 
\usepackage{amsmath} 
\usepackage{amssymb}  
\usepackage{algorithmic,algorithm}
\usepackage{url}
\usepackage{romannum}

\title{\LARGE \bf
Memorable Maps: A Framework for Re-defining Places in Visual Place Recognition
}

\author{Mubariz Zaffar$^{1}$, Shoaib Ehsan$^{1}$, Michael Milford$^{2}$ and Klaus McDonald-Maier$^{1}$
\thanks{$^{1}$Authors are with the School of Computer Science and Electronic Engineering,
        University of Essex, CO4 3SQ, United Kingdom
        {\tt\small mz18963@essex.ac.uk, sehsan@essex.ac.uk, kdm@essex.ac.uk}}%
\thanks{$^{2}$Michael Milford is with the School of Electrical Engineering and Computer Science, Queensland University of Technology, Brisbane, QLD 4000, Australia
        {\tt\small michael.milford@qut.edu.au}}%
\thanks{This work is supported by the UK Engineering and Physical Sciences Research Council through grants EP/R02572X/1 and EP/P017487/1.}
}

\begin{document}

\maketitle
\thispagestyle{empty}
\pagestyle{empty}

\begin{abstract}
   This paper presents a cognition-inspired agnostic framework for building a map for Visual Place Recognition. This framework draws inspiration from human-memorability, utilizes the traditional image entropy concept and computes the static content in an image; thereby presenting a tri-folded criterion to assess the ‘memorability’ of an image for visual place recognition. A dataset namely ‘ESSEX3IN1’ is created, composed of highly confusing images from indoor, outdoor and natural scenes for analysis. When used in conjunction with state-of-the-art visual place recognition methods, the proposed framework provides significant performance boost to these techniques, as evidenced by results on ESSEX3IN1 and other public datasets. 
\end{abstract}


\section{Introduction}

Visual Place Recognition (VPR) is a well-defined but highly challenging module of a Visual-SLAM (Simultaneous Localization and Mapping) based autonomous system \cite{1}. It represents the ability of a robot to `remember' a previously visited place in the world map and thus subsequently generating a belief about the robot's location in the world. VPR is mostly used in combination with metric SLAM techniques to perform loop closure \cite{18}. Recent advances in SLAM research can be broken down into semantic mapping (surveyed in \cite{kostavelis2015semantic}) and Visual Place Recognition (surveyed in \cite{1}), where the latter can be annexed into the former \cite{kostavelis2015semantic}. This specific work concerns metric, topological and topometric maps having single/multiple images as nodes (landmarks) of the map.   

\begin{figure}[t]
\begin{center}
\includegraphics[width=1\linewidth]{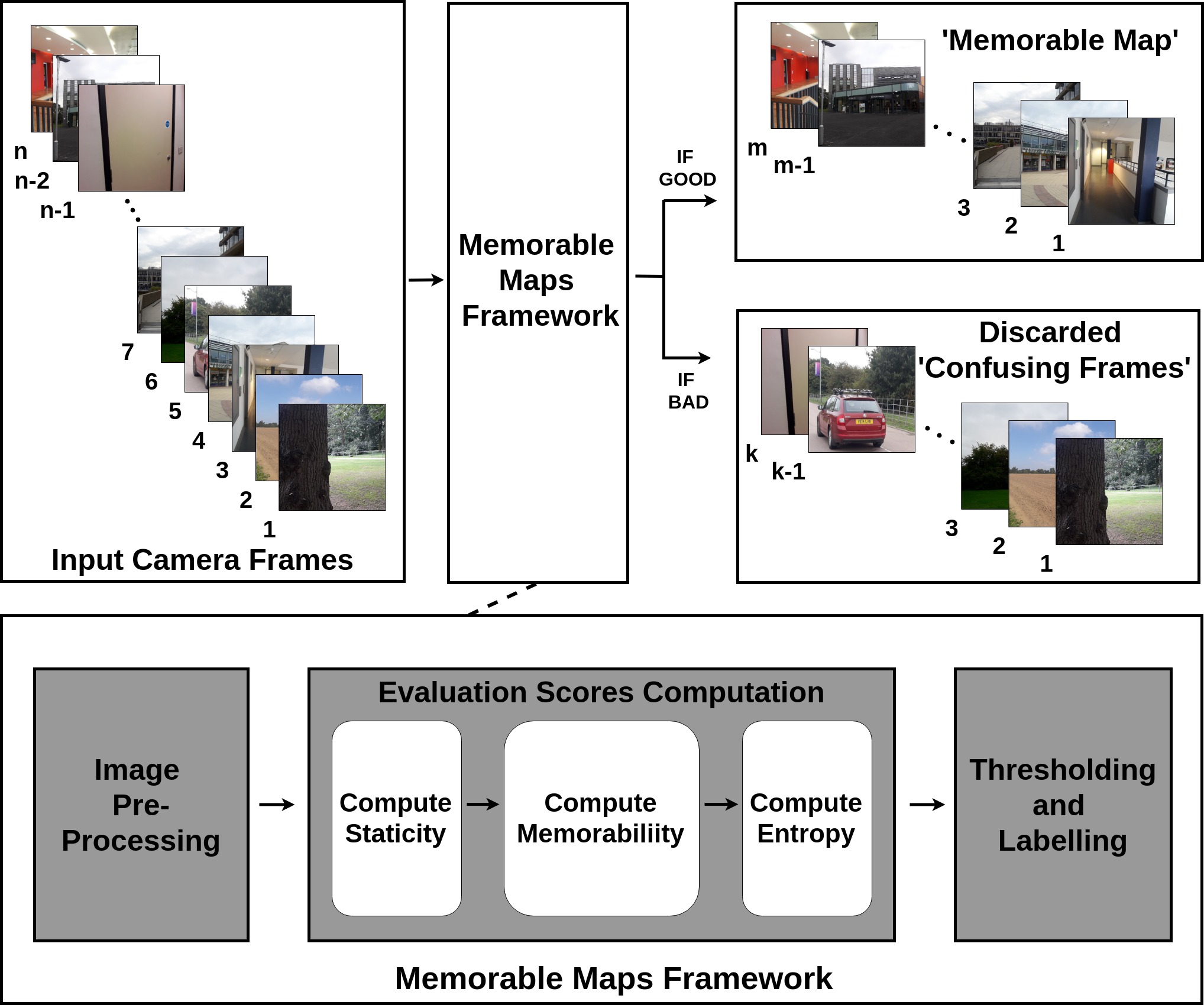}
\end{center}
\caption{A block-level overview of the proposed memorable-maps framework. Each image is evaluated for its memorability, staticity and entropy before insertion into `memorable-map'.}
\label{Fig:Memorable_Maps_Framework_IO}
\end{figure}

Traditionally, for visual place recognition, `Places' have been selected/sampled based on time-interval \cite{3}, distance \cite{12} or distinctiveness \cite{13} in different approaches. These approaches are discussed in depth in the next section. Most of these methods attempt to reduce the size of robot's map and do not quantify if a sampled/sub-sampled image is a good representation of a place; thereby has a greater chance of matching upon revisiting. The quality of image selection mechanism restricts the performance of a VPR system, both in the short-term and long-term. Due to limited number of images being stored in the map, it is critical to select those images that can be matched successfully upon repeated traversal-the motivation for this research. 

In this work, we look at image selection from a semantic point-of-view and draw inspiration from images memorable to a human-cognition system. We use a Convolutional Neural Network \cite{2} to compute the memorability of an incoming camera frame. However, while objects like cars, pedestrians and vehicles in an image are subjectively-memorable; they are intrinsically not good for place recognition as such dynamic objects are rarely re-observed. We thus perform object detection to compute the staticity of an image and mask memorability of dynamic content. In addition to being memorable and static, an image has to be content-rich for which we calculate the entropy map. 

The contribution of this work is a semantically coherent framework (Fig.  \ref{Fig:Memorable_Maps_Framework_IO}) that filters an input image through a tri-folded criterion. Hence, ensuring that every image to be inserted against a place in robot's map is a good representation of the said place and highly recognizable. To analyze the effectiveness of this framework, we created a dataset `ESSEX3IN1' from indoor, outdoor and natural environments. Unlike existing VPR datasets, ESSEX3IN1 mimics a robot exploring an environment instead of traditional path-following and is thus composed of highly confusing images from all three environments. We show how these confusing images lead to poor performance of current visual place recognition systems. The final results show the effectiveness of proposed framework in segregating these `confusing' images from `good' images, thereby increasing VPR precision and reducing database size. We also evaluate our framework on other public VPR datasets to show that this performance enhancement can be generalized. Due to its agnostic nature, any VPR technique can obtain a performance boost by stacking the presented framework as an additional layer in the VPR pipeline. 

The remainder of the paper is organized as follows. In Section \Romannum{2}, a comprehensive literature review regarding Visual Place Recognition state-of-the-art is presented with focus on image selection and semantic-mapping. Section \Romannum{3} presents the motivation, design and implementation details of the `memorable-maps' framework developed in this work. Section \Romannum{4} is dedicated to the experimental setup for evaluating and analyzing state-of-the-art VPR techniques with and without proposed framework. Following-up on Section \Romannum{4}, Section \Romannum{5} puts forth the results/analysis obtained by combining memorable-maps and contemporary VPR techniques on multiple public datasets and ESSEX3IN1. Finally, conclusions and future directions are presented in Section \Romannum{6}.    

\section{Related Work}
VPR and SLAM have seen major developments through different cognitive, intuitive or semantic approaches to the problem. A comprehensive review of these techniques is performed by Lowry et al. \cite{1}. An earlier work on probablistic implementaion of SLAM in visual-appearance domain, called `FAB-MAP', is presented by Cummins et al. \cite{cummins2008fab}. This work was combined in \cite{glover2010fab} with a biologically inspired SLAM technique `RAT-SLAM' \cite{milford2004ratslam}, mimicking the Rat's hippocampus. Milford et al.  \cite{6} utilize sequence of images instead of individual frames to successfully match previously visited places under significant environmental variations. Similar to other fields, Convolutional Neural Networks (CNNs) have been a game-changer for VPR. The application of CNN for VPR was first studied by Chen et al. \cite{chen2014convolutional}. Authors in \cite{20} trained two dedicated Neural Networks for VPR on Specific Places Dataset (SPED) containing images from different seasons and times of day. Unlike previous implementations where image descriptors were manually formed from CNN layer activations, Arandjelovic et al. \cite{arandjelovic2016netvlad} trained a new VLAD (Vector of Locally Aggregated Descriptors) layer for an end-to-end CNN-based-VPR.  For images containing repetitive structures, Torii et al. \cite{torii2013visual} proposed a robust mechanism for collecting visual words into descriptors. Synthetic views are utilized for enhanced illumination invariant VPR in \cite{torii201524}, which shows that highly condition variant images can still be matched if they are from the same viewpoint. In \cite{chen2017only}, authors try to extract local features from convolutional layers corresponding to salient Regions of Interest (ROI), thus providing significant viewpoint invariance. State-of-the-art performance is shown by authors in \cite{21}, by combining VLAD descriptors with ROI-extraction to show immunity to appearance and viewpoint variation.        

Traditionally, places have been described by camera frames, where a place is selected from multiple video frames based on either time-step, distance or distinctiveness. Most of the VPR datasets \cite{3} \cite{4} \cite{5} \cite{6} \cite{7} \cite{8} \cite{9} \cite{10} are time-based, as frames are selected given a fixed FPS (frames per second) rate of a video camera. However, time-based place selection assumes a constant non-zero speed of the robotic platform and is thus impractical in many situations. To cater for variable speed, distance-based frame selection is used where a frame is picked every few metres to represent a new place \cite {11} \cite{12}. Both time- and distance-based approaches lead to huge database sizes and frequently sample visually identical frames as different places; thus leading to inaccuracies and impracticality for long-term autonomy.  

Different research works have tried to overcome these intrinsic limitations of image sampling by proposing image selection based on visual distinctiveness. Chapoulie et al. \cite{13} use a customised algorithm that detects change point for segmentation between different topological places in both indoor and outdoor scenes.  Image sequence partitioning for creating sparse topological maps is presented by Korrapati et al. \cite{14}, where sequences of images are divided into nodes/places using four descriptors namely GIST, Optical Flow, Local Feature Mapping and Common-Important Words. In \cite{15}, a thematic approach is adapted to evaluate the novelty of an incoming image by co-relating it with the redundancy of visual features/topics. Bayesian surprise is adapted with immunity to sensor type, for extracting landmarks to create a sparse topological map in \cite{16}. Online topic modeling with visual surprise calculation is done by Girdhar et al.  \cite{17} for under-water explorations. An incremental unsupervised place discovery scheme is adopted by Murphy et al. \cite{18} which fuses information over time to find visually distinct places.

Semantic mapping techniques for summarizing a robot's experience are surveyed by Kostavelis et al.  \cite{kostavelis2015semantic}. Authors in \cite{girdhar2010online} present both offline and online solutions for finding images that best summarize a given sequence. The score for every incoming image is related to the difference of posterior distribution from prior distribution using bayesian surprise or set theoretic surprise.  In \cite{paul2014visual}, coresets are used to pre-cluster input image stream and then topic-based image representation is used followed with graph-based incremental clustering. A place detection scheme is proposed by Karaoguz et al.  \cite{karaoguz2014reliable} based on bubble-space representation. A new place is checked for informativeness based on surface deformation and variance in a time-window of coherent images. The authors in \cite{demir2018automated} use region proposals in spatio-temporal context instead of low-level features to represent input frames and then based on region-adjacency-graph detect visually distinct places. A human-augmented change point detection scheme is presented by Topp et al. \cite{topp2008detecting} where a change stimuli could either be pointed out by the robot or its operator. The authors propose the change as a structural ambiguity, which can be pointed out either by the robot or a human operator during a guided tour. Detection of change point is also targeted by Ranganathan \cite{ranganathan2013detecting} with a Bayesian probablistic model. One common element to all these works is that they focus on map compression, video segmentation or experience summarization, but do not discuss if the resulting compressed/summarized map is actually composed of good matchable images of places. These methods define the distinctive nature of images based on their visual difference from previously seen images. Resultingly, such visually different images may come from grassy plains, natural scenery, dynamic objects or low-textured places leading to poor VPR performance. Drawing inspiration from the said, we in this work, define distinctiveness based on a place's memorability (cognitive), static-content and information-richness leading to highly matchable compressed maps dubbed as `memorable-maps'.

In addition to the above, we discuss two works having similar motivation to our approach. The interesting work by Hartmann et al. \cite{hartmann2014predicting} proposes a random forest classifier of 5 decision trees trained on a dataset of 455 outdoor images. The objective of this random forest is to find keypoints in an image with low matchability and subsequently discarding them. This technique is computationally intensive in comparison to our methodology as we compute a single matchability (memorability) score against an image instead of scores against each image keypoint. Moreover, in VPR, features coming from dynamic objects and low-textured scenes are usually not re-observable/matchable (as shown later in our paper) but have not been examined in \cite{hartmann2014predicting}. Although vegetation is considered to belong to non-matchable category, results show features coming from trees as being classified as matchable in \cite{hartmann2014predicting}; which usually in VPR contribute negatively to the distinctiveness of a place (as shown in Fig.  \ref{Fig:Mismatched_Low_Memorability_Images}). More recently, a CNN able to classify input frames as stable/unstable is trained by Dymczyk et al. \cite{dymczyk2016will} for long term visual place recognition. Similar to \cite{hartmann2014predicting}, this work also proposes that vegetation in outdoor scenes is not suitable, but does not consider outdoor dynamic objects like cars, pedestrians, animals etc. Also, informativeness of stable frames in terms of extracted features and predicted matchability is not inspected given that walls are selected as stable elements of an image. Therefore, to the best of author's knowledge, our work combines for the first time all three of these criterion namely memorability, staticity and entropy to create memorable-maps. We show utility of our proposed framework by reporting results on multiple public datasets. The agonstic nature of our framework is presented by using multiple state-of-the-art VPR techniques in combination with memorable-maps.   

\section{Methodology}
\begin{figure}[h]
\begin{center}
\includegraphics[width=1.0\linewidth]{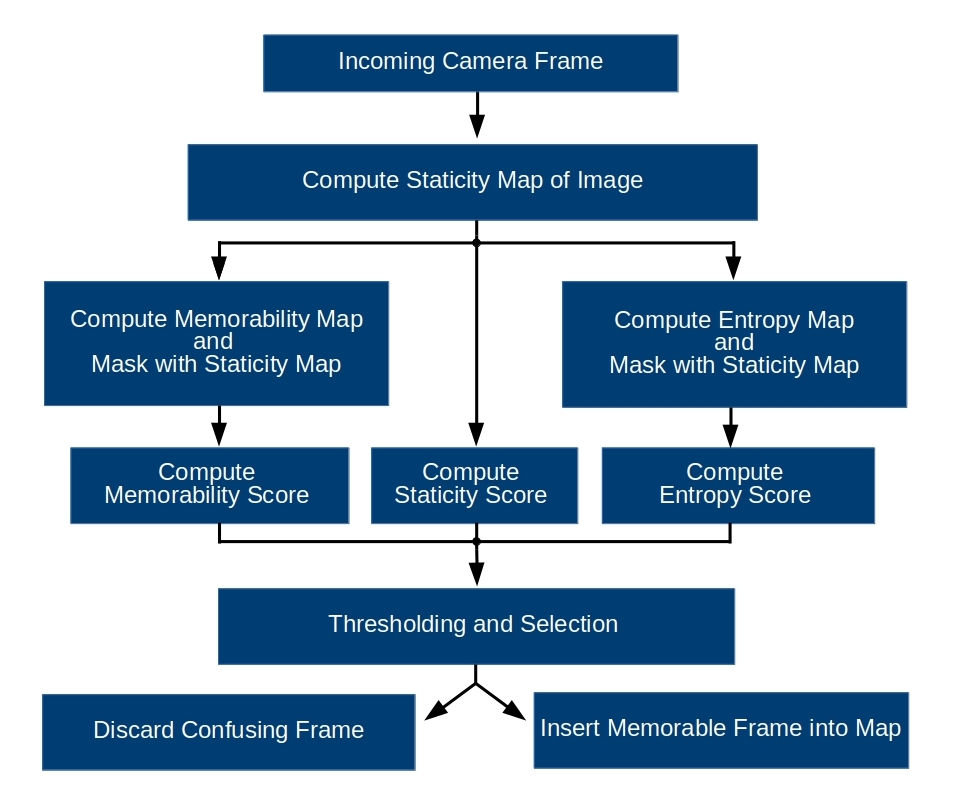}
\end{center}
\caption{Flowchart of the proposed framework. Each frame first goes through staticity-map computation. Next, we compute the frame's memorability-map and entropy-map which are masked with staticity-map. Finally, we calculate the scores from all three image-maps, threshold and discard confusing frames.}
\label{Fig:Framework_Block_Diagram}
\end{figure}

This section presents in depth the framework developed in our work. A sub-section is dedicated to each of the three criterion (i.e. memorability, staticity and entropy) adopted by the framework. Flowchart depicting the working of memorable-maps framework is shown in Fig.  \ref{Fig:Framework_Block_Diagram}.  

For the purpose of evaluation and analysis, we have used AMOS-Net \cite{20}, Hybrid-Net \cite{20} and Region-VLAD \cite{21} as our VPR techniques throughout this paper. The details of these techniques are given in Section \ref{VPR_techniques}.  
\subsection{Memorability}

\subsubsection{Why Memorability?} \label{WhyMemorability?}

\begin{figure}[h]
\begin{center}
\includegraphics[width=1\linewidth, height=0.8\linewidth]{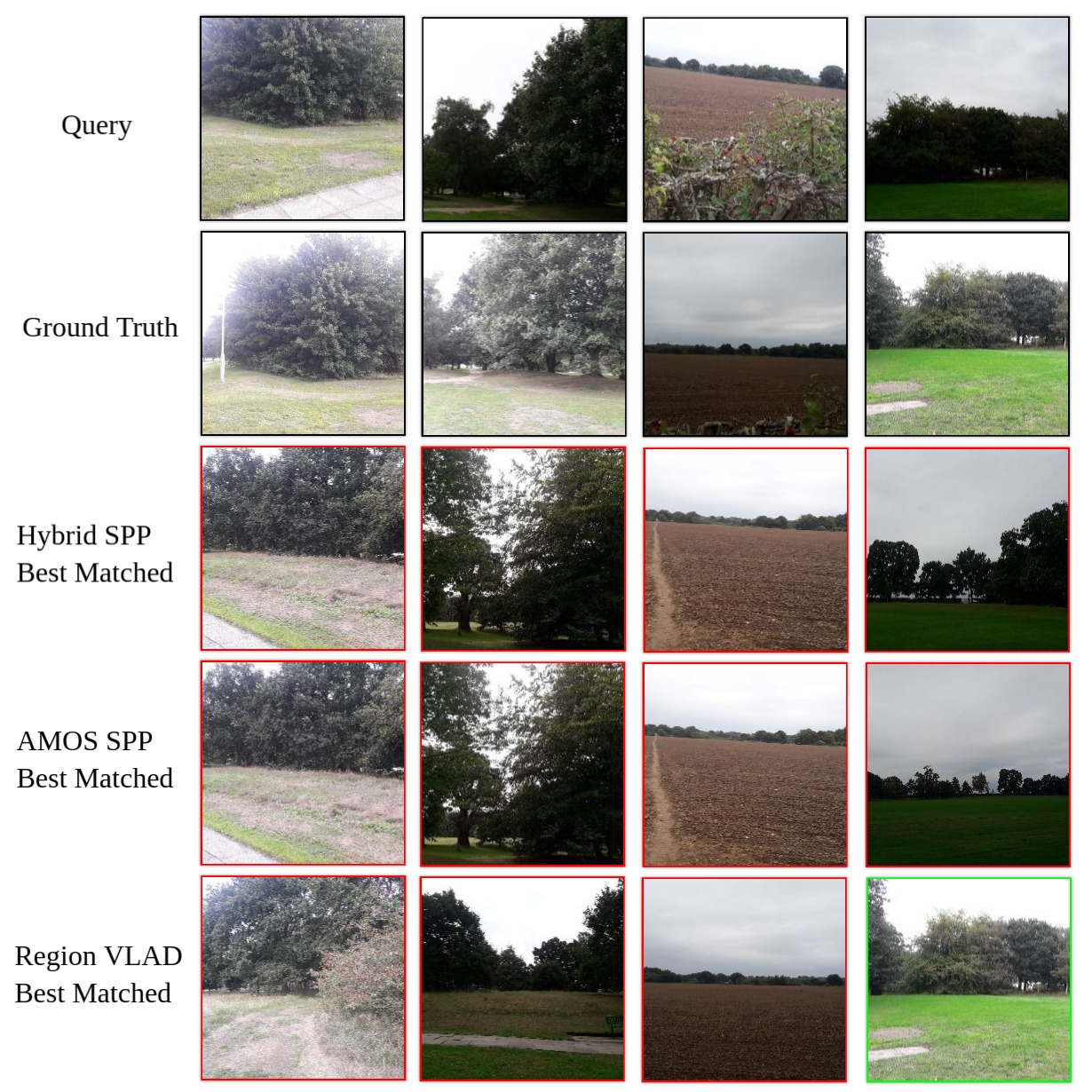}
\end{center}
\caption{Natural Places mismatched by VPR methods due to confusing and indistinguishable features coming from trees, grass and plains.}
\label{Fig:Mismatched_Low_Memorability_Images}
\end{figure}

The human-cognition system is powerful in evaluating what images are useful to be stored in the brain's memory fragments \cite{brady2008visual} \cite{konkle2010scene}. We usually remember concrete structures like buildings, streets, squares etc. However, more natural scenes like fields, forests, grassy plains and far out sceneries are less memorable. This `memorability' concept is also intuitive as it is easy to confuse different natural scenes with each other compared to concrete structures. In Fig. \ref{Fig:Mismatched_Low_Memorability_Images}, we show how such natural scenes are mismatched by state-of-the-art VPR systems leading to false-positives. 

\subsubsection{Memorability Implementation} 
Inspired from human-memorability, we apply the work done originally for marketing and advertising in \cite{2} to VPR problem. A Convolutional Neural Network namely Hybrid-CNN is fine-tuned on LaMem dataset (introduced in \cite{2}). This LaMem dataset is composed of 60,000 images covering multiple scenarios ranging from natural scenery, indoor scenes, outdoor scenes and distinctive objects. The ground-truth human-memorability is computed for each of these images using an interactive game played by multiple human subjects. Images are shown to players in a sequence and are repeated after a random interval where a human has to identify/recall a previously seen image. Resultingly, the output of such fine-tuned Hybrid-CNN is a human-memorablity score $m$ for each input image in the range of $0-1$; with $m=1$ being the most memorable. 

The CNN input layer size is set to $W1 \times H1$ where $W1=227$, $H1=227$. We re-size every incoming image as follows.
\begin{equation*}
\begin{aligned}
RescaledImage &= Rescale(Image, W2, H2) \\
where; \;\;\;\; W2&=5\times W1 \\ 
H2&=5\times H1
\end{aligned}
\end{equation*}
We then split this rescaled-image into 25 non-overlapping crops of $227 \times 227$ each and sequentially feed them as inputs to CNN. This in-turn gives us the memorability matrix $M$ as shown below.

\[M=
\begin{bmatrix}
    m_{11} & m_{12} & \dots & m_{15} \\
    m_{21} & m_{22} & \dots & m_{25} \\
    \vdots & \vdots	& \ddots & \vdots \\
    m_{51} & m_{52} & \dots & m_{55}
\end{bmatrix}
\]

Where, $m_{ij}$ is the memorability of each $227 \times 227$ cropped image. To create a memorability-map, we rescale the matrix $M$ from $5 \times 5$ to $W2 \times H2$ with bilinear interpolation. An example memorability-map overlayed on an image is shown in Fig.  \ref{Fig:Framework_Criterion_Maps}. It can be seen (in Fig. \ref{Fig:Framework_Criterion_Maps}) that vegetation, natural scenery and trees are identified as less-memorable which is consistent with our motivation in Section \ref{WhyMemorability?}. However, for human-cognition (and therefore for the fine-tuned Hybrid-CNN) objects such as faces, animals and vehicles are memorable. But, such dynamic objects are not re-observable and therefore not good for VPR; we cater for this in the following sub-section.  
 
\subsection{Staticity}

\subsubsection{Why Staticity?}
\begin{figure}[h]
\begin{center}
\includegraphics[width=1\linewidth, height=0.8\linewidth]{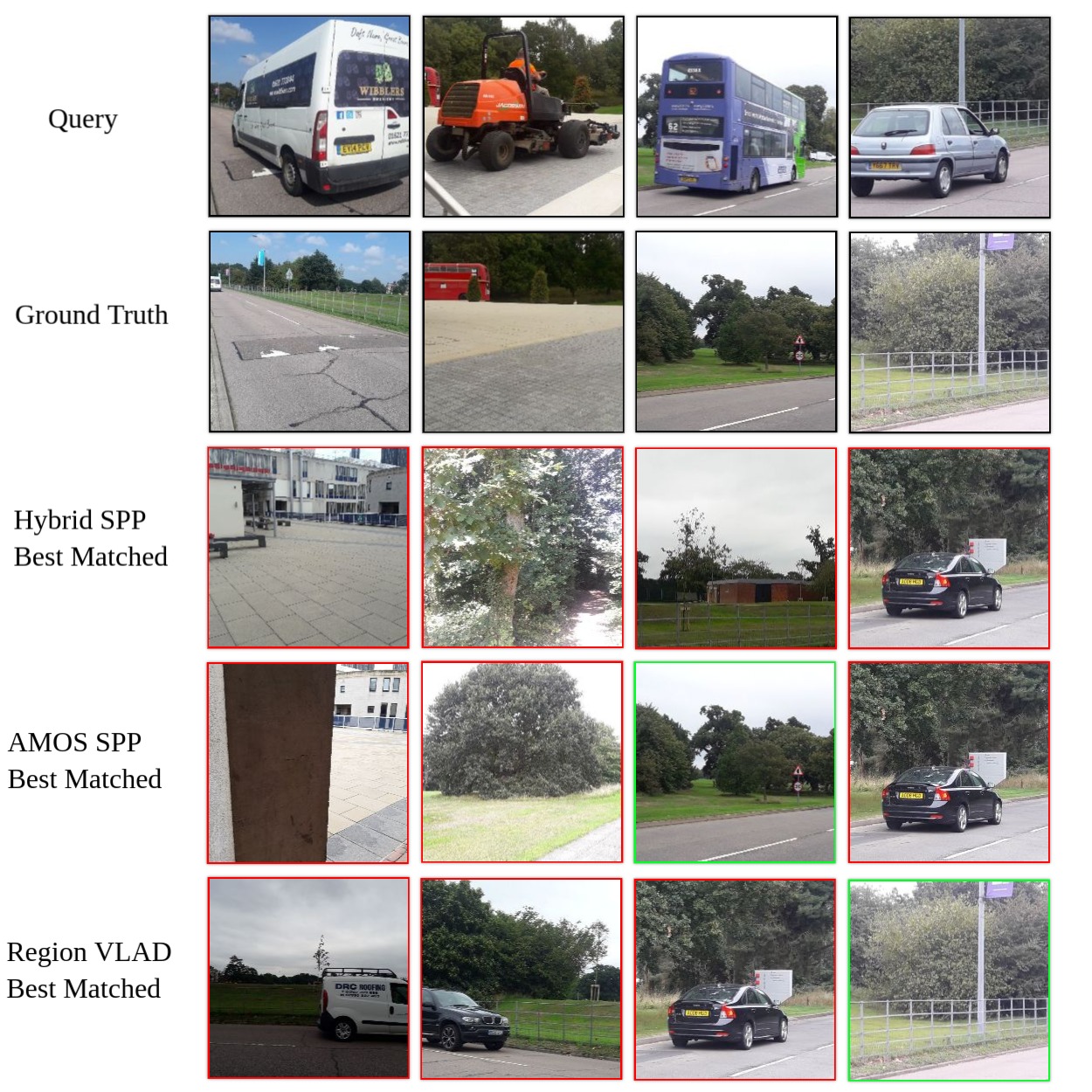}
\end{center}
\caption{Dynamic places mismatched by state-of-the-art VPR systems. Features coming from vehicles are not re-observable in addition to the occlusion caused by them in different scenes.}
\label{Fig:Mismatched_Low_Staticity_Images}
\end{figure}

The previous sub-section shows how memorability is a good evaluation criterion for a camera frame to be used in VPR. However, one limitation is the fact that objects like cars, pedestrians, buses, animals and bicycles in an image are all classified as highly memorable but are not re-observable (for VPR problem). Resultingly, images that may be memorable but have high dynamic content will fail to match upon repeated traversal. Fig.  \ref{Fig:Mismatched_Low_Staticity_Images} shows some of these images mismatched by VPR techniques \cite{20} \cite{21}. 
 
\subsubsection{Staticity Implementation}
To cater for highly dynamic images, we perform image segmentation into static and dynamic pixels. We re-size all input images to $W2 \times H2$. We use an object detector \cite{redmon2017yolo9000} that can detect $80$ different classes of objects in an image. Out of these $80$ classes, $21$ correspond to highly-dynamic, commonly-observed objects. These dynamic objects include cars, pedestrians, buses, trucks, animals etc. We, therefore, only consider proposals of bounding boxes coming from objects of interest, i.e., dynamic objects. 

The object detector also requires a confidence metric (between zero and one) to give bounding box proposals. We selected this confidence metric as $0.55$, which performs well for objects that occupy at least $5\%$ of the total image pixels. Objects occupying less than $5\%$ of the total pixels are ignored by our framework. Hence, by not detecting these low significance objects, the selected confidence metric works in coherence with our criterion. After acquiring bounding boxes which are usually overlapping, we use the algorithm given below to compute staticity-map. \\

\begin{algorithm} [H]
\renewcommand\thealgorithm{}
\caption{Computing Staticity Map}
\begin{algorithmic}
\STATE $Initially \; All\_Image\_Pixels \gets Static $ \\ 

\FORALL{$Pixels \; in \; Image$}
\FORALL{$Bounding\_Boxes \; in \; Image$}
\IF{$Current\_Pixel \; lies \; within \; Bounding\_Box$}
\STATE {$Current\_Pixel \gets Dynamic $ \\
		$Break \; Loop $
}
\ENDIF
\ENDFOR
\ENDFOR

\end{algorithmic}
\addtocounter{algorithm}{-1}
\end{algorithm}

Since the staticity map is computed for each pixel in the Image, it can be represented as a staticity-matrix $S$ of size $W2 \times H2$ as below.   

\[S=
\begin{bmatrix}
    s_{11} & s_{12} & \dots & s_{1W_2} \\
    s_{21} & s_{22} & \dots & s_{2W_2} \\
    \vdots & \vdots	& \ddots & \vdots \\
    s_{H_2W_2} & s_{H_2W_2} & \dots & s_{H_2W_2}
\end{bmatrix}
\]

\begin{equation*}
\begin{aligned}
where; \; \; \; \; \; \{s_{ij} \in \mathbb{Z}_2 \; \: | \; \: \mathbb{Z}_2 &= [0,1] \}
 \\
s_{ij}=1 | Pixel&=Static \\
s_{ij}=0 | Pixel&=Dynamic 
\end{aligned}
\end{equation*}
Fig. \ref{Fig:Framework_Criterion_Maps} shows the typical staticity-map computed in our framework. However, although an image containing low-textured scenes (walls/door/pillars) can be classified as concrete (memorable) and static but it does not have distinguishable features, and hence is not distinct. We  accommodate this limitation in the following sub-section.

\subsection{Entropy}

\subsubsection{Why Entropy?}
\begin{figure}[h]
\begin{center}
\includegraphics[width=1\linewidth, height=0.8\linewidth ]{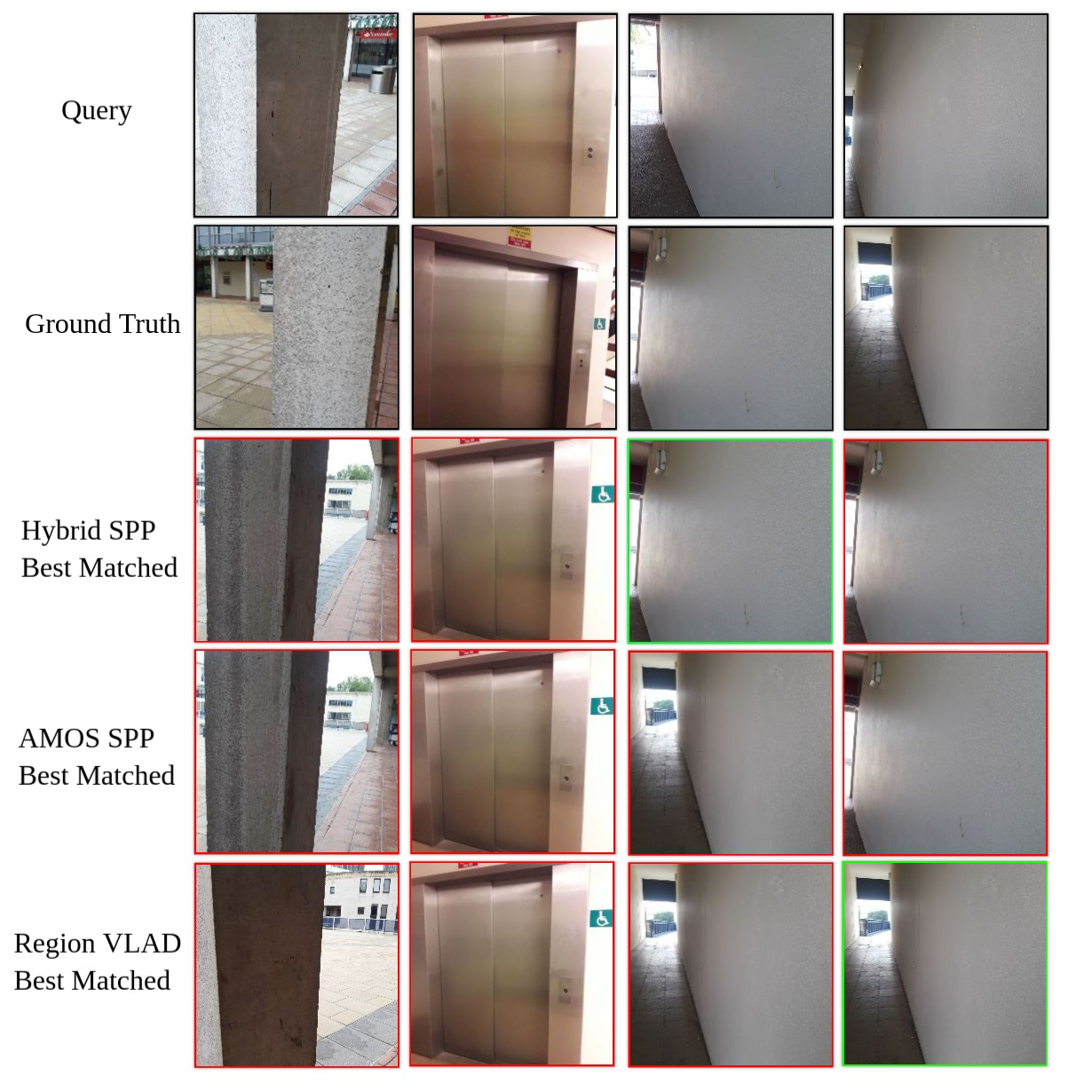}
\end{center}
\caption{Low-entropy places mismatched by state-of-the-art VPR methods can be commonly observed in indoor robot navigation datasets. Alongwith intrinsically less-lnformative images of doors/walls, static occlusion can also lead to poorly defined places.}
\label{Fig:Mismatched_Low_Entropy_Images}
\end{figure}

An input camera frame containing a room/lift door is commonly observed by a robot navigating indoors. Such a frame is classified as memorable and static, but has little to no information differentiating it from other doors in the building, thus leading to false positives. The same can be extended to any other frame with occlusion resulting from walls, pillars etc. Examples of such confusing frames are shown in Fig. \ref{Fig:Mismatched_Low_Entropy_Images}.

\subsubsection{Entropy Implementation}
To avoid less informative or occluded frames, we evaluate the information content of an image by computing its local entropy. This local entropy corresponds to the number of bits required to encode the local gray-scale distribution in an image. We use a circular window of 5 pixels radius as our neighbourhood to get the entropy map of an incoming camera frame against each pixel. The total number of histogram bins used for entropy computation are $256$ corresponding to $0-255$ gray-scale intensity values. The generic algorithm for computing this entropy map is shown below and adopted from \cite{van2014scikit}.

\begin{algorithm} [H]
\renewcommand\thealgorithm{}
\caption{Computing Entropy Map}
\begin{algorithmic}
\STATE $Create \; a \; Histogram \; of \; 256 \; Bins $ 

\FORALL{$Local \; Neighbourhoods \; in \; Image$}
\FORALL{$Pixels \; in \; Current \; Neighbourhood$}
\IF{$Current\_Pixel \; lies \; in \; BinX$}
\STATE {$Items\_in\_BinX=Items\_in\_BinX + 1 $ }
\ENDIF 
\ENDFOR \\
$Local \_ Entropy=log_2\:(No.\:of\: Filled\:Histogram\:Bins)$ \\
$Clear \: all \: Histogram \: Bins$
\ENDFOR

\end{algorithmic}
\addtocounter{algorithm}{-1}
\end{algorithm}

\begin{figure*}[h]
\begin{center}
\includegraphics[width=0.95\linewidth, height=0.35\linewidth]{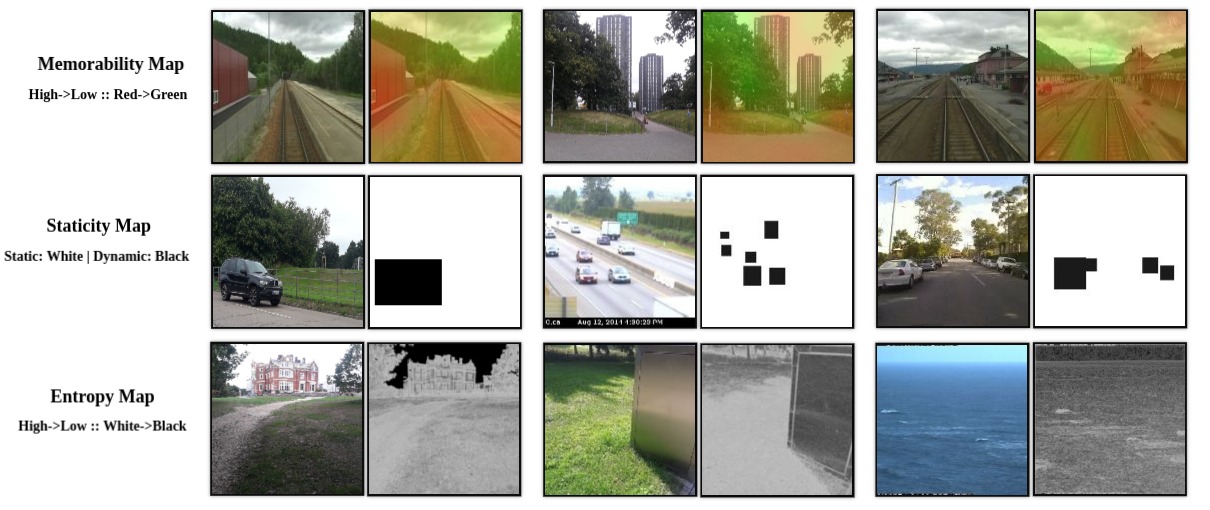}
\end{center}
\caption{The three types of image maps created by proposed framework for evaluating the content of an input image. Concrete structures like buildings and roads are memorable in comparison to grassy plains and trees [Top]. Cars, pedestrians and other dynamic objects are detected and evaluated for the amount (approximate) of pixels they occupy [Middle]. Uniform scenes, large occlusions or sky portions have low-entropy compared to feature rich structures [Bottom].}
\label{Fig:Framework_Criterion_Maps}
\end{figure*}

This algorithm gives us an entropy map represented as matrix $E$ of size $W2 \times H2$. The maximum value of entropy is computed from equation (1) and equals 8. Fig.  \ref{Fig:Framework_Criterion_Maps} shows an example entropy map computed in our framework. 
\begin{equation}
Max\:Entropy = log_2\:(No.\:of\:Histogram\:Bins)
\end{equation}

\[E=
\begin{bmatrix}
    e_{11} & e_{12} & \dots & e_{1W_2} \\
    e_{21} & e_{22} & \dots & e_{2W_2} \\
    \vdots & \vdots	& \ddots & \vdots \\
    e_{H_2W_2} & e_{H_2W_2} & \dots & e_{H_2W_2}
\end{bmatrix}
\]

\begin{equation*}
\begin{aligned} 
where; \; \; \; \; \; \{ e_{ij} \in K \; \: | \; \: K \subseteq  \mathbb{R} \wedge K=\{ 0, \dots, 8 \}\}
\end{aligned}
\end{equation*}

\subsection{Computing Scores and Thresholding}

After acquiring all three maps of an image, we mask memorability map and entropy map with staticity map. This ensures that our decision to select an image based upon memorability and entropy is immune to the information coming from dynamic objects.    

Next, we compute the memorability score ($MS$) of an image as the average value of memorability map and compare it with a memorability threshold ($MT$), to evaluate if this image/frame is memorable enough for use in VPR.  

Secondly, we compute the percentage of static pixels in our staticity map to get a staticity score ($SS$). This is then contrasted with staticity threshold ($ST$) to decide if an incoming frame has enough static content to be inserted into the map. 

Thirdly, we calculate the average value of entropy map and scale it with the maximum value of entropy to get the percentage of information content. This percentage dubbed as the entropy score ($ES$), is compared with the entropy threshold ($ET$) to settle if an input frame has enough information. 

Finally, we use a tri-input AND criterion to select images that are memorable, static and information-rich to be inserted into the map.

\begin{algorithm} [H]
\renewcommand\thealgorithm{}
\caption{Image Selection For Memorable Map}
\begin{algorithmic}
\FORALL{$Incoming \; Images$}
\STATE {$\; \; \; \; \; \; \; \; \;Compute \; All \; Three \; Image \; Maps$}

\begin{flalign*} 
 \; \; \; \; \; \; \; \; \; MS =  \frac {1} {W2 \times H2 } \sum_{i,j=1,1}^{W2,H2} m_{ij} &&
\end{flalign*}
\noindent
\begin{flalign*} 
 \; \; \; \; \; \; \; \; \; SS =  \frac {1} {W2 \times H2 } \sum_{i,j=1,1}^{W2,H2} s_{ij} &&
\end{flalign*}
\noindent
\begin{flalign*} 
\; \; \; \; \; \; \; \; \; ES =  \frac {1} {W2 \times H2 \times 8} \sum_{i,j=1,1}^{W2,H2} e_{ij} &&
\end{flalign*}

\IF{$MS\geq MT \; \& \; SS\geq ST \; \& \; ES\geq ET$}
\STATE {$Insert \; into \; Map$ }
\ELSE
\STATE {$Dicard \; Image$}
\ENDIF 
\ENDFOR 
\end{algorithmic}
\addtocounter{algorithm}{-1}
\end{algorithm}

\section{Experimental Setup}

This section discusses the datasets, VPR techniques and evaluation metric used in our analysis. We present a new dataset ESSEX3IN1, which will be publicly available. Additionally, we briefly discuss three pre-existing public datasets used for reporting our framework's  performance. The VPR techniques used for our results and analysis are then summarized. We utlize area-under-the Precision-Recall curve (AUC) which is a well-established performance metric for VPR techniques \cite{chen2018learning} \cite{chen2017only} \cite{arandjelovic2016netvlad} \cite{chen2014convolutional} \cite{21}.

\subsection{Evaluation Datasets} \label{Evaluation_Datasets}
  
\subsubsection{ESSEX3IN1 Dataset}

Most of the Visual Place Recognition datasets have been created from a pre-planned path traversal. Thus, these datasets do not contain confusing images that an exploration robot may come across. Also, these datasets focus on a single type of environment either indoor or outdoor. To evaluate and challenge our framework, we have created a new dataset ESSEX3IN1 which is composed of images from indoor, outdoor and natural scenes. 

\begin{figure}[h]
\begin{center}
\includegraphics[width=1\linewidth, height=0.7\linewidth]{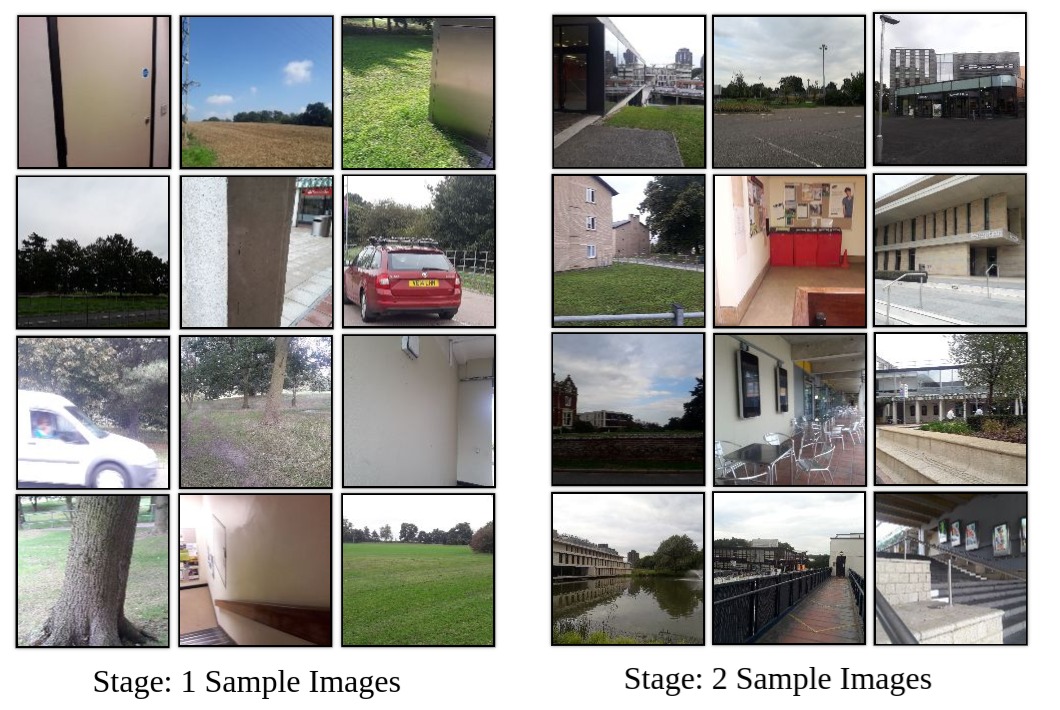}
\end{center}
\caption{Sample images from ESSEX3IN1 dataset. The first stage [on the left hand side] images contain occlusions, dynamic objects, information-less frames and non-memorable content like plains, natural scenery, vegetation and trees. In contrast, the second stage [on the right hand side] contains semantically identifiable and distinguishable images of various places from University of Essex (Colchester campus).}
\label{Fig:ESSEX3IN1_Samples}
\end{figure}

The dataset was created in two stages at the University of Essex (Colchester Campus) and contains 210 query images and 210 reference images with viewpoint variations. In the first stage, the objective was to take images from all sorts of environments that were either `confusing' or didn't qualify the human-definition of `Place'. Two-third of the images in ESSEX3IN1 are from this first stage. The second stage, consists of images that were not confusing and could be defined as places. One-third of the total images are from this second stage. Some images from these stages are shown in Fig. \ref{Fig:ESSEX3IN1_Samples}. 

It is important to note that none of these images were used in tuning our three thresholds and were not seen prior by the proposed framework. The collection of dataset in this two-staged manner was useful for analysis in Section \ref{ResultsandAnalysis}. 

\subsubsection{Nordland Dataset}

\begin{figure}[h]
\begin{center}
\includegraphics[width=0.95\linewidth, height=0.45\linewidth]{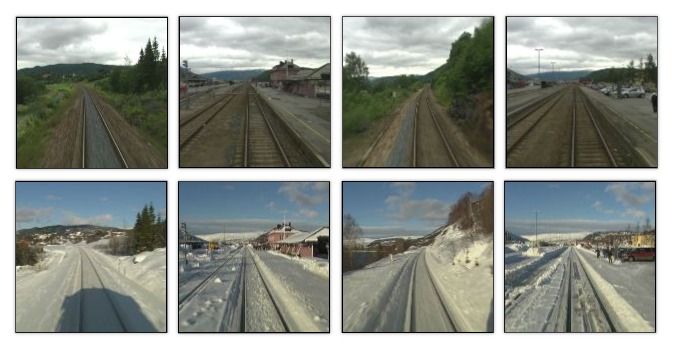}
\end{center}
\caption{Sample images from Nordland dataset. Top row consists of query images from summer traversal while bottom row consists of reference images from winter traversal.}
\label{Fig:Nordland_Samples}
\end{figure}

The Nordland dataset \cite{7}  comprises of a train journey through Norway and is collected in four different seasons with frame-to-frame ground-truth correspondence. We use a subset of this dataset which consists of 1622 query images and 1622 reference images. The query images are from the traversal performed in summer where as the reference images are from winter. Although this dataset does not provide any viewpoint variation, but has significant conditional variation. Some sample images from Nordland dataset are shown in Fig. \ref{Fig:Nordland_Samples}.  A retrieved image $n$ is considered true-positive if the original ground-truth is between $n-1$ to $n+1$ i.e. each query image has $3$ ground-truth references.   

\subsubsection{St. Lucia Dataset}
The St. Lucia dataset was first introduced in \cite{glover2010fab}. It was recorded in the surroundings of University of Queensland's St. Lucia campus during multiple times of the day. This dataset consists of moderate viewpoint and illumination variation as shown in Fig.  \ref{Fig:Stlucia_Samples}. The dataset also contains dynamic objects and scene variation. The ground-truth is derived manually from GPS data such that each query image has three reference images as true-positives. 

\begin{figure}[h]
\begin{center}
\includegraphics[width=0.95\linewidth, height=0.45\linewidth]{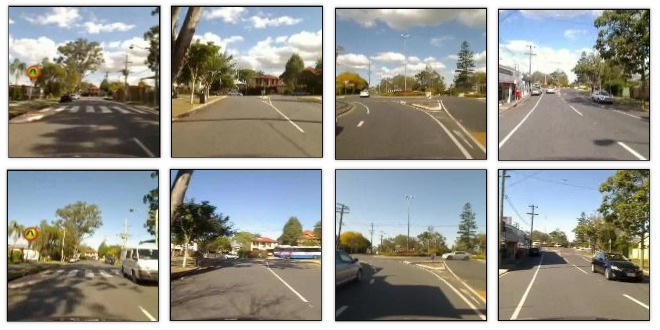}
\end{center}
\caption{Sample images from St. Lucia dataset. Top row consists of query images from 14:10 traversal while bottom row consists of reference images from 08:45 traversal.}
\label{Fig:Stlucia_Samples}
\end{figure}

\subsubsection{SPEDTest Dataset}
The SPEDTest dataset was introduced in \cite{chen2014convolutional} and is a sub-set of the original Specific Places Dataset \cite{20}. It consists of 607 query images coming from a variety of scenes and environments. Frame-to-frame correspondence is available as the ground-truth. Samples images from SPEDTest are shown in Fig. \ref{Fig:Stlucia_Samples}.  

\begin{figure}[h]
\begin{center}
\includegraphics[width=0.95\linewidth, height=0.45\linewidth]{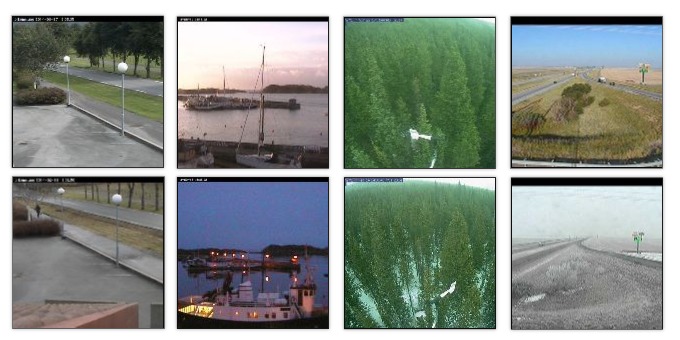}
\end{center}
\caption{Sample images from SPEDTest dataset. Top row consists of query images and the bottom is corresponding ground-truth reference image.}
\label{Fig:Stlucia_Samples}
\end{figure}

\subsection{VPR Techniques} \label{VPR_techniques}
We have used three state-of-the-art VPR techniques (namely AMOS-SPP, Hybrid-SPP and Region-VLAD) \cite{20} \cite{21} that have shown promising results in recent research. AMOS-Net is a modified Caffe-Net \cite{krizhevsky2012imagenet} with all parameters trained on SPED dataset \cite{20}. Hybrid-Net is another modified version of Caffe-Net with weights for top 5 convolutional layers initialized from Caffe-Net \cite{krizhevsky2012imagenet}. We have used Spatial Pyramidal Pooling as a feature descriptor for both AMOS-Net and Hybrid-Net since it shows excellent results as compared to other feature encoding methods. Features are extracted from `conv5' layer in case of both Amos-Net and Hybrid-Net. The third VPR technique, Region-VLAD, uses features extracted from selected/interesting regions of an AlexNet pre-trained on Places365 dataset. Vector-of-Locally-Aggregated-Descriptors \cite{jegou2010aggregating} is subsequently used for encoding the extracted features. In case of Region-VLAD, we use features from `conv4', number of regions-of-interest as $400$ and a visual dictionary size of $128$. 

\subsection{Evaluation Metric}
For evaluating the performance of different VPR techniques AUC has been repeatedly used by the research community. AUC acts as a good metric to assess the performance of a system based on true-positives, false-positives and false-negatives. However, AUC scores may differ depending on the method employed for computing the area. In our work, we use equation (\ref{AUC_equation}) to compute the AUC of a Precision-Recall curve.

\begin{equation} \label{AUC_equation}
AUC = \sum_{i=1}^{N-1} \frac {(p_i + p_{i+1})} {2} \times (r_{i+1} - r_i)
\end{equation}

\begin{equation*}
\begin{aligned}
where; \; \; N = No. \; of \; Query \; Images
 \\
p_i=Precision \; at \; point \; i \\
r_i= Recall \; at \; point \; i 
\end{aligned}
\end{equation*}

This paper ensures consistency and fair comparison of AUC scores for different VPR method on all datasets by computing and reporting results only using equation (\ref{AUC_equation}). 

\section{Results and Analysis} \label{ResultsandAnalysis}

This section presents the results and analysis in a sequential manner. We first show that images collected from the first stage of ESSEX3IN1 actually lead to poor performance of VPR systems and are not good for insertion into a topological-map. Secondly, we show the segregation performance of proposed framework on these `confusing' images and `good' images. Thirdly, we present the AUC improvement of different VPR systems when plugged with our framework on all datasets discussed in sub-section \ref{Evaluation_Datasets}. This is followed-up with a sub-section dedicated to qualitative analysis showing sample images selected and discarded from all datasets. We then highlight the contribution of each framework criterion and its effect on VPR performance by sweeping within possible range. And finally, we show how such a framework leads to reduced map size and place matching time.

\subsection{Contemporary VPR Systems on ESSEX3IN1 Stage 1}

\begin{figure}[h]
\begin{center}
\includegraphics[width=1\linewidth, height=0.7\linewidth]{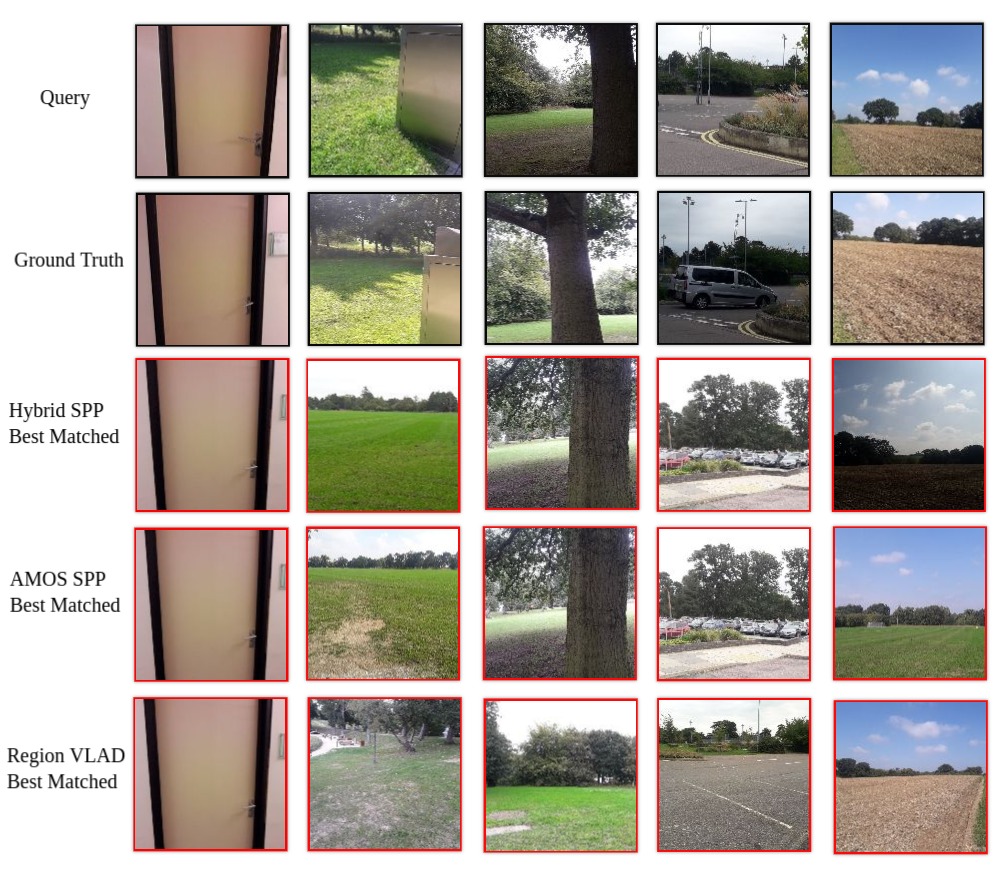}
\end{center}
\caption{VPR false positives upon evaluation on ESSEX3IN1 stage: 1. Images with cars, trees and natural scenes are mismatched. Additionally, images with low information and memorability are almost indistinguishable for even human cognition.}
\label{Fig:VPR_False_Positives_Stage1}
\end{figure}

The majority of VPR false positives against ESSEX3IN1 are from the first stage of dataset collection. This is due to the confusing images of fields, trees, doors, cars etc. Some of these false positives are shown in Fig. \ref{Fig:VPR_False_Positives_Stage1}.

\begin{figure}[h]
\begin{center}
\includegraphics[width=1\linewidth]{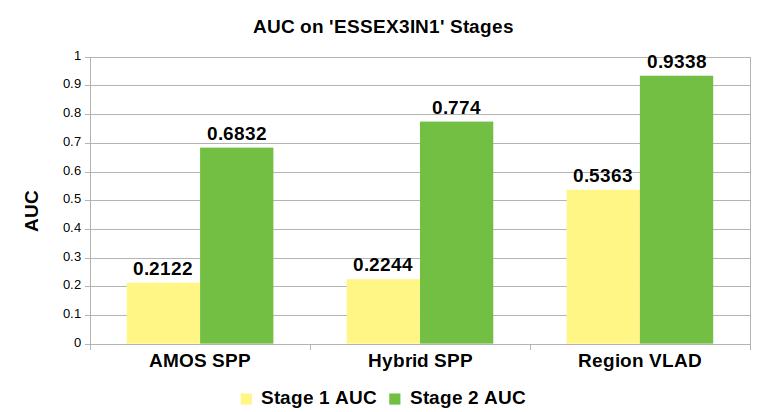}
\end{center}
\caption{Separate evaluation of VPR methods on each of ESSEX3IN1 stages reveals the challenge faced by contemporary VPR techniques for matching low-entropy, low-memorability and dynamic images.}
\label{Fig:AUC_ON_ESSEX3IN1_Stages}
\end{figure}

We show the AUC performance of VPR systems separately on Stage 1 and Stage 2 in Fig.  \ref{Fig:AUC_ON_ESSEX3IN1_Stages}.

\subsection{Segregation Performance of Proposed Framework}
\label{Segregation Performance of Proposed Framework}
\begin{figure}[h]
\begin{center}
\includegraphics[width=1\linewidth]{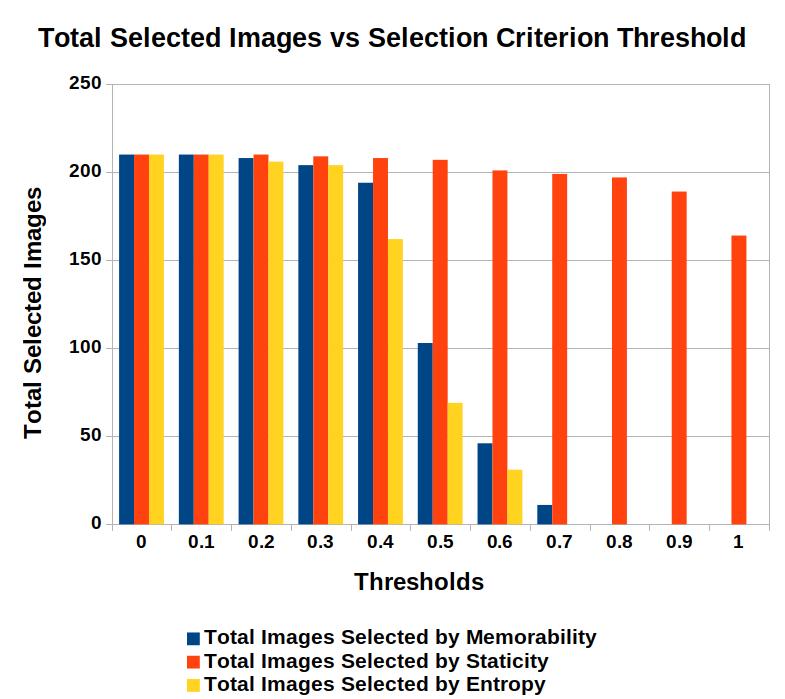}
\end{center}
\caption{The decrease in total selected images as each selection criterion is increased, can be observed. Majority of the ET/MT based image selection is done between $0.4-0.7$. Purely static images (without vehicles, human and animals) exist in the dataset which is why $ST=1$ does not reduce map size to zero.}
\label{Fig:SelectedImagesvsFrameworkCriteria}
\end{figure}

For this sub-section, we apply the proposed memorable-maps framework on complete and randomized ESSEX3IN1 dataset. We use the below thresholds to segregate and discard `confusing' images from `good' images. 
\begin{equation*}
\text{Memorability-threshold} = 0.5 
\end{equation*}
\begin{equation*}
\text{Staticity-threshold} = 0.6 
\end{equation*}
\begin{equation*}
\text{Entropy-threshold} = 0.4 
\end{equation*}
These thresholds were selected from our analysis on pre-existing public VPR datasets. Increasing these thresholds reduces the number of images inserted into the memorable-map. This is shown in Fig. \ref{Fig:SelectedImagesvsFrameworkCriteria} by varying each threshold from $0-1$ while setting the other two equal to $0$. 

\begin{figure}[h]
\begin{center}
\includegraphics[width=1\linewidth]{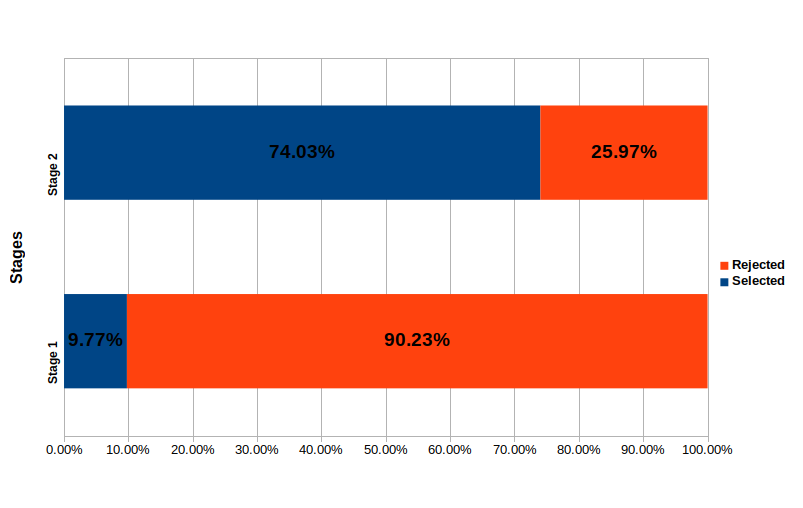}
\end{center}
\caption{The objective of memorable-maps framework is to sample good frames and discard confusing frames. This objective achievement is presented by showing the contribution in memorable-map from each ESSEX3IN1 stage.}
\label{Fig:ImagesSelectedFromEachStage}
\end{figure}

The new database created by presented framework consists of memorable, static and informative images, thus dubbed as a memorable-map. We show [in Fig. \ref{Fig:ImagesSelectedFromEachStage}] how many of the total images selected by presented framework are from which stage of the dataset.     

\subsection{AUC Improvement of VPR Systems}

By selecting images that are memorable, static and have a higher entropy, the memorable-maps framework gives performance boost to state-of-the-art VPR techniques. Here, we use fixed thresholds, as in previous sub-section \ref{Segregation Performance of Proposed Framework}, but an AUC sweep across these thresholds is presented later in sub-section \ref{AUC_Sweep}. AUC evaluation is performed on the entire (both stages combined randomly) ESSEX3IN1 dataset along with the three public VPR datasets. It is important to note that bad/confusing images found by our framework are not removed from the reference database when evaluating AUC, but are treated as true negatives. This ensures that AUC boost reported here is not due to reduction of reference database size.

\begin{figure}[h]
\begin{center}
\includegraphics[width=1\linewidth]{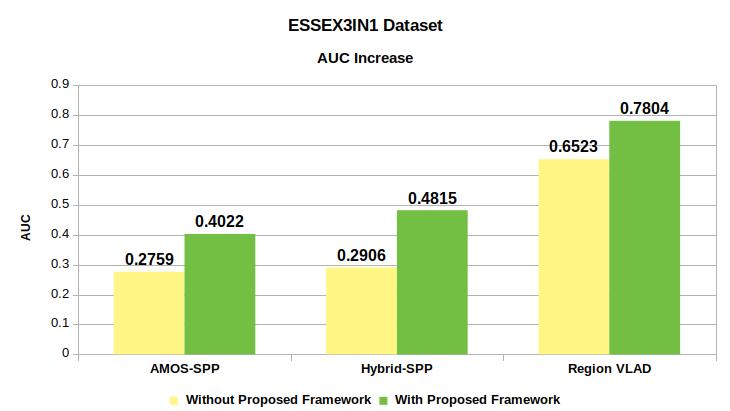}
\end{center}
\caption{Increase in AUC by using proposed framework in combination with VPR techniques on complete, randomly-shuffled ESSEX3IN1 dataset is presented. Reference database size remained same for all AUC evaluations.}
\label{Fig:ESSEX3IN1_AUC_BOOST}
\end{figure}

\begin{figure}[h]
\begin{center}
\includegraphics[width=1\linewidth]{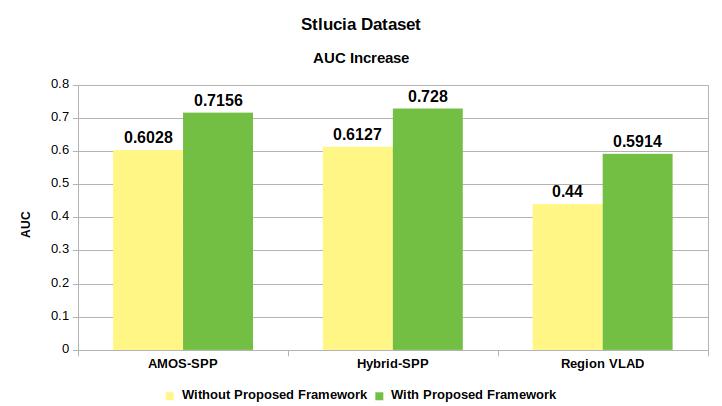}
\end{center}
\caption{Increase in AUC by using proposed framework in combination with VPR techniques on St. Lucia dataset is presented. Reference database size remained same for all AUC evaluations.}
\label{Fig:Stlucia_AUC_BOOST}
\end{figure}

\begin{figure}[h]
\begin{center}
\includegraphics[width=1\linewidth]{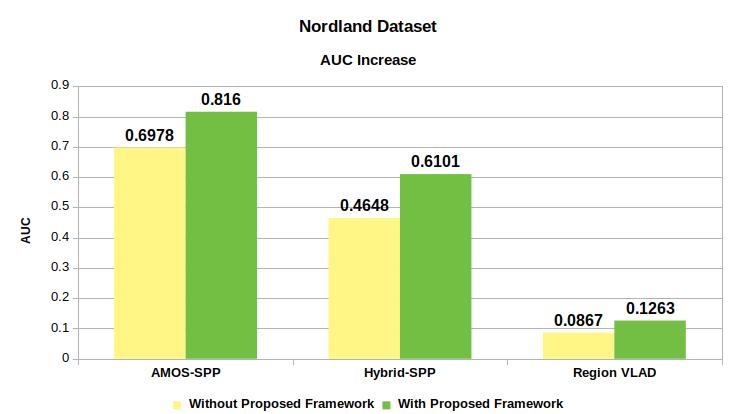}
\end{center}
\caption{Increase in AUC by using proposed framework in combination with VPR techniques on Nordland dataset is presented. Region-VLAD doesn't perform well on this dataset however the AUC increase is consistent. Reference database size remained same for all AUC evaluations.}
\label{Fig:Nordland_AUC_BOOST}
\end{figure}

\begin{figure}[h]
\begin{center}
\includegraphics[width=1\linewidth]{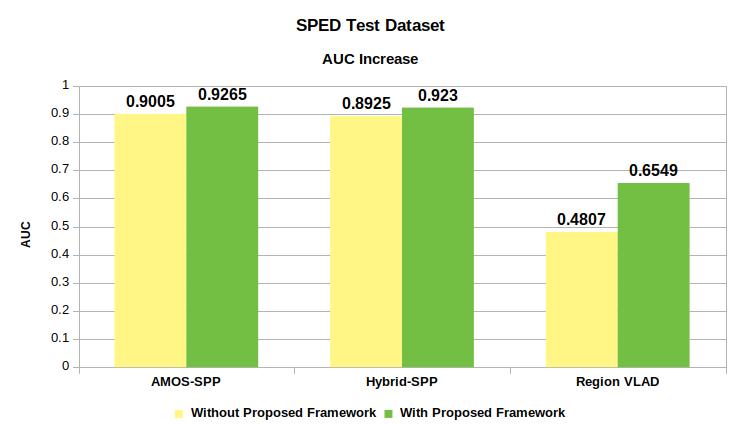}
\end{center}
\caption{Increase in AUC by using proposed framework in combination with VPR techniques on SPEDTest dataset is presented. AMOSNet and HybridNet being trained on the SPED dataset perform very well in comparison to Region-VLAD. Reference database size remained same for all AUC evaluations.}
\label{Fig:SPEDTest_AUC_BOOST}
\end{figure}

Fig. \ref{Fig:ESSEX3IN1_AUC_BOOST}, \ref{Fig:Stlucia_AUC_BOOST}, \ref{Fig:Nordland_AUC_BOOST} and \ref{Fig:SPEDTest_AUC_BOOST} depict the AUC increase by employing our framework on ESSEX3IN1, St. Lucia, Nordland and SPEDTest dataset respectively. We use the same values for MT, ST and ET as in sub-section \ref{Segregation Performance of Proposed Framework} for ESSEX3IN1, Nordland and SPEDTest dataset. However, for St. Lucia we reduce each of the three selection thresholds by $0.05$ to get a non-zero map size.    

\subsection{Selected vs Discarded Images}
\begin{figure}[h]
\begin{center}
\includegraphics[width=1\linewidth]{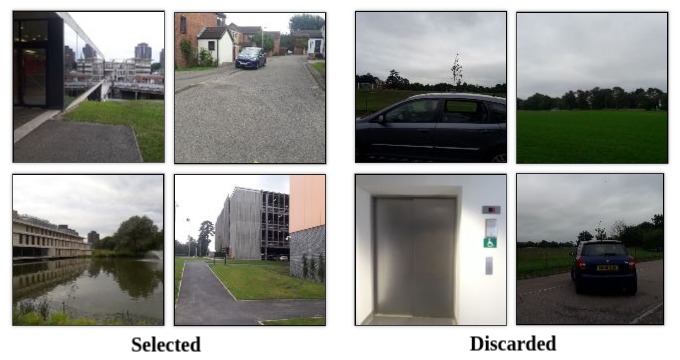}
\end{center}
\caption{Images selected and discarded by the memorable-maps framework from complete ESSEX3IN1 dataset are presented. Selected images are pre-dominantly of buildings with distinctive patterns and are largely static. Discarded images consist of far out natural scenes, dynamic objects or have low-entropy.}
\label{Fig:essex3in1_selectedvsdiscarded}
\end{figure}

\begin{figure}[h]
\begin{center}
\includegraphics[width=1\linewidth]{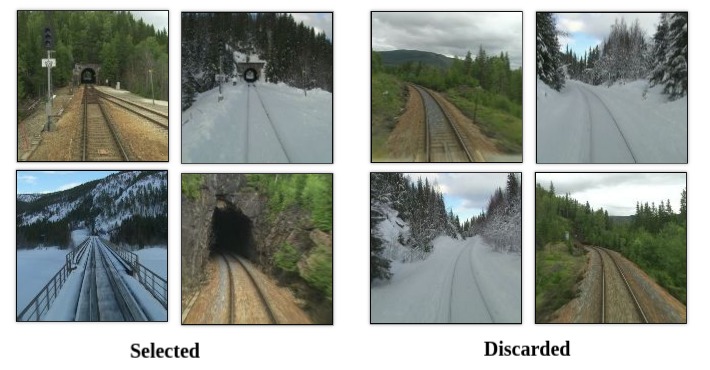}
\end{center}
\caption{Images selected and discarded by the memorable-maps framework from Nordland dataset are presented. Selected images consist of either appearing tunnels or bridges which contribute to their distinctiveness. Discarded images consist of vegetation or have low information. Staticity does not play any role in this dataset due to absence of dynamic objects.}
\label{Fig:nordland_selectedvsdiscarded}
\end{figure}

\begin{figure}[h]
\begin{center}
\includegraphics[width=1\linewidth]{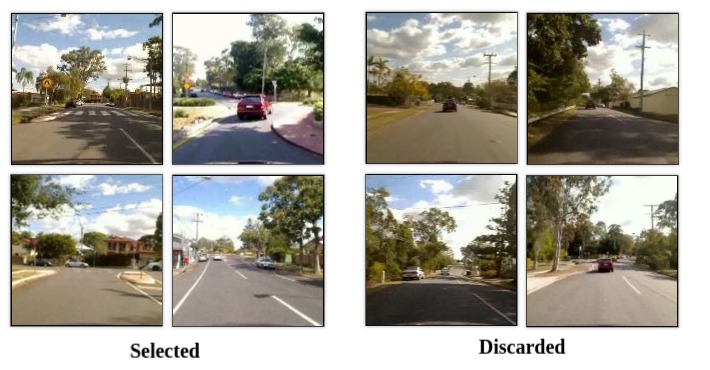}
\end{center}
\caption{Images selected and discarded by the memorable-maps framework from St. Lucia dataset are presented. Selected images contain road signs, squares and houses. On the other hand, discarded images comprise of far out road scenes with trees and large portions of sky.}
\label{Fig:stlucia_selectedvsdiscarded}
\end{figure}

\begin{figure}[h]
\begin{center}
\includegraphics[width=1\linewidth]{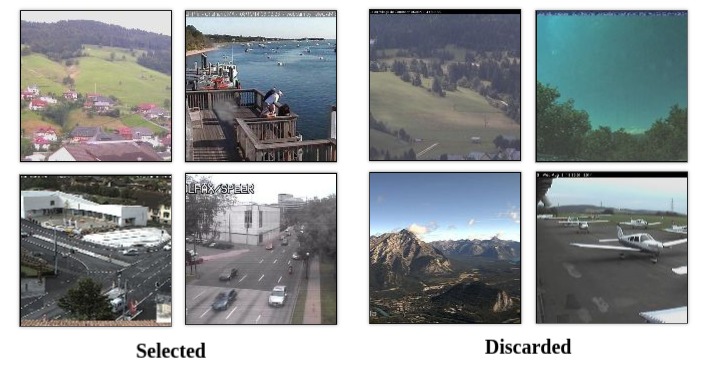}
\end{center}
\caption{Images selected and discarded by the memorable-maps framework from SPEDTest dataset are presented. Selected images are from CCTVs covering buildings or distinctive locations. Discarded images consist of far out natural scenes, dynamic objects or have low entropy.}
\label{Fig:spedtest_selectedvsdiscarded}
\end{figure}

In this sub-section, we show some images from all four datasets that were selected or discarded by the memorable-maps framework. This gives a qualitative insight into the working of our framework in different environments/datasets. Since the framework evaluates both the query images and reference images, the images in Fig. \ref{Fig:essex3in1_selectedvsdiscarded}, \ref{Fig:nordland_selectedvsdiscarded}, \ref{Fig:stlucia_selectedvsdiscarded} and \ref{Fig:spedtest_selectedvsdiscarded} are impartial to such distinction. 

\subsection{Criterion Contribution Analysis}
\begin{figure}[h]
\begin{center}
\includegraphics[width=1\linewidth]{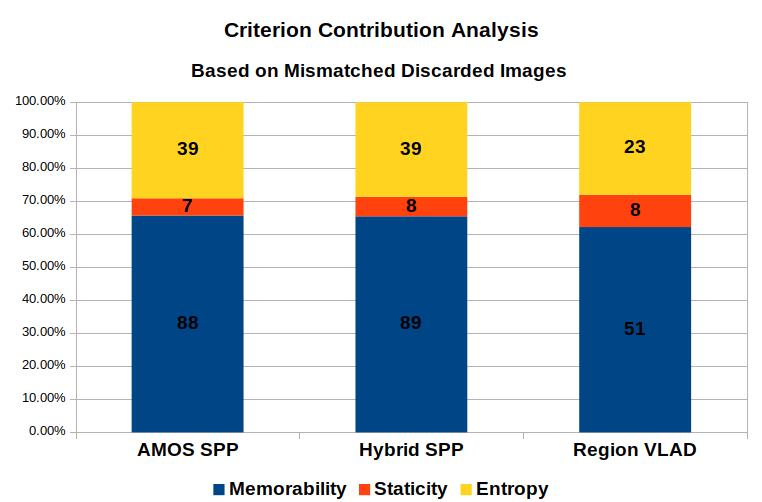}
\end{center}
\caption{Percentage contribution of each criterion into AUC increase is shown for ESSEX3IN1. This contribution is directly linked with the type of environment being explored by a robot. In a highly dynamic environment, the contribution of staticity will be more significant than suggested by this chart and such.}
\label{Fig:Criterion_Contribution_Analysis}
\end{figure}

Each criterion in the memorable-maps framework contributes to AUC boost. This subsection is dedicated to giving an insight into this indiviudal contribution. We use ESSEX3IN1 for this purpose as it contains confusing images from all three (memorability, staticity and entropy) paradigms. For our AUC evaluation on ESSEX3IN1, we show the contribution of each criterion in Fig. \ref{Fig:Criterion_Contribution_Analysis}. The analysis is performed based on the number of images that were mismatched by a VPR technique and were also discarded by atleast one of the memorable-maps framework criterion.

While Fig. \ref{Fig:Criterion_Contribution_Analysis} suggests that each of the three criterion namely memorability, staticity and entropy are useful; the \% contribution is linked to (and can vary with) the number of non-memorable, dynamic and information-less images in the dataset. (See Fig. \ref{Fig:SelectedImagesvsFrameworkCriteria})

\subsection{AUC Sweep Across Framework Thresholds} \label{AUC_Sweep}

In this subsection, we present the variation in Visual Place Recognition performance with stricting framework criteria on ESSEX3IN1. We sweep each of the three criterion from 0-1 (Step size: 0.1) while keeping the other two inactive (i.e. set equal to zero). The data points for memorability and entropy thresholds have an upper-bound after which the total number of selected images equals to $0$ (refer Fig. \ref{Fig:SelectedImagesvsFrameworkCriteria}).   

\begin{figure*}[h]
\begin{center}
\includegraphics[width=0.85\linewidth, height=0.5\linewidth]{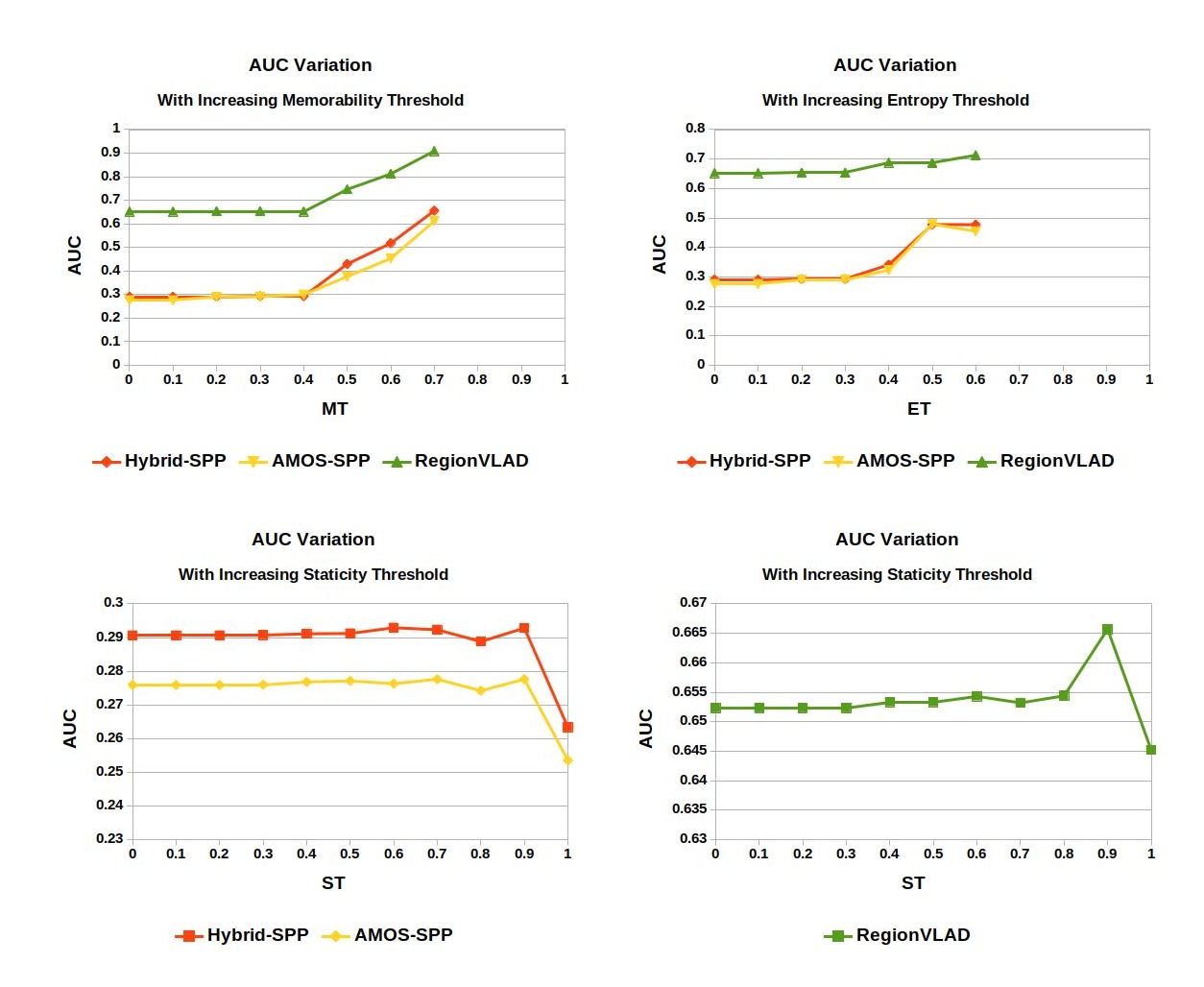}
\end{center}
\caption{Variation in VPR AUC performance by changing each of the memorable-maps framework thresholds within their full range on ESSEX3IN1 is presented. Memorability and entropy continously increase AUC until the total number of selected images equals to zero; suggesting that images with higher memorability and entropy are well-matched by VPR methods. On the contrary, since images with low dynamic content should/can still be matched, variation in staticity threshold does not lead to a continous AUC increase. AUC change with ST is not at the same scale as MT/ET so it is shown separately.}
\label{Fig:AUC_Sweep_Combined}
\end{figure*}

Fig. \ref{Fig:AUC_Sweep_Combined} shows that increasing entropy-threshold and memorability-threshold increases the AUC Performance for all three VPR techniques and follows a direct-relationship. On the other hand, the variation in AUC with increasing staticity-threshold follows a different trend. Firstly, the increase in AUC with ST is comparitively lower compared to MT/ET; which is due to the less number of dynamic images in the dataset compared to non-memorable and low-entropy images. Secondly, the variation in AUC with ST for Region-VLAD is higher compared to AMOS-SPP/Hybrid-SPP. We associate this with the fact that AMOS-SPP/Hybrid-SPP have been trained on SPED (Specific Places Dataset) and discourage features coming from vehicles. While our analysis/results reveal that Region-VLAD extracts and positively matches features coming from cars in different places (See Fig.  \ref{Fig:Mismatched_Low_Staticity_Images}). Thirdly, there is an evident decrease in AUC as ST goes above $0.9$. This decrease is expected as images with very low dynamic content can still be matched by contemporary VPR-techniques and discarding such images leads to the observed decline in VPR-performance. The presented trends give a general idea for setting thresholds, thus to maintain a good balance between VPR performance and a salient representation of the world in a metric/topological/topo-metric map.      
  
\subsection{Reduced Map Size and Computational Time}
\begin{figure}[h]
\begin{center}
\includegraphics[width=0.9\linewidth, height=0.5\linewidth]{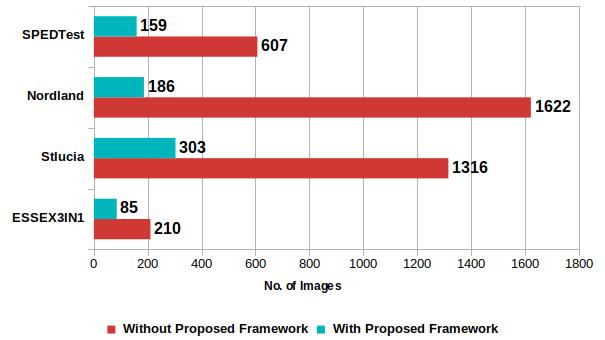}
\end{center}
\caption{Reduction in topological map size given similar or better VPR results is of prime importance for an autonomous robot to efficiently map/explore an environment. As depicted here, memorable-maps framework intrinsically reduces map size while giving AUC boost to contemporary VPR systems.}
\label{Fig:Map_Size_Reduction}
\end{figure}

In addition to increase in AUC, the developed framework helps in reducing the robot's map size which has been the motivation for semantic mapping research reviewed in this paper. This size-reduction also leads to reduction in computational overhead for VPR. The reduction in map size for thresholds presented in Section \ref{Segregation Performance of Proposed Framework} is shown in Fig. \ref{Fig:Map_Size_Reduction}. The computational performance is reported by calculating the time required to match a query image with all the reference images (having pre-computed feature descriptors) in both a conventional map and a memorable map. This offline matching time is elaborated in Table 1, where a memorable map having lesser number of reference images (please see Fig. \ref{Fig:Map_Size_Reduction}) achieves better matching time.  

\begin{table*}[t]
\centering
\caption{Matching Time Reduction}
\begin{tabular}{|c|c|c|c|c|c|c|}

\hline
System Specifications & \multicolumn{6}{|c|}{Intel(R) Xeon(R) Gold 6134 CPU @ 3.20GHz, 64GB Physical Memory} \\ 
\hline 
Framework & \multicolumn{3}{|c|}{Without Memorable Maps} & \multicolumn{3}{|c|}{With Memorable Maps} \\ 
\hline 
VPR Method & AMOS-SPP & Hybrid-SPP & Region-VLAD & AMOS-SPP & Hybrid-SPP & Region-VLAD\\ 
\hline 
ESSEX3IN1 (sec) & 10.2 & 9.9 & 0.14 & 4.1 & 3.9 & 0.05 \\ 
\hline 
Nordland (sec) & 78.7 & 76.4 & 1.1 & 9.1 & 8.7 & 0.12 \\ 
\hline 
St. Lucia (sec) & 63.9 & 62.1 & 0.88 & 14.7 & 14.2 & 0.21 \\ 
\hline  
SPEDTest (sec) & 29.5 & 28.6 & 0.41 & 7.7 & 7.5 & 0.11 \\ 
\hline 
\end{tabular} 
\end{table*}

\section{Conclusion and Future Work}
We proposed a cognition-inspired generalized framework for creating `memorable-maps'. This framework evaluates an incoming camera frame for its memorability, staticity and entropy to decide a frame's insertion into the robot's topological map. By using `ESSEX3IN1', we show how images that are confusing for human-beings are also mismatched by contemporary VPR systems. The application of proposed framework in detecting these confusing images and subsequently improving VPR performance is presented. We generalize the applicability of our framework by reporting results on multiple public datasets. Due to its agnostic nature, memorable-maps framework can be plugged into any VPR technique giving performance boost. 

While presented thresholds are suitable for different indoor, outdoor and natural environments, they are not illumination invariant. In future work, it will be useful to integrate \cite{22} to this work, thus making these thresholds as illumination-dependent variables. This framework coupled with different VPR techniques also enables the creation of a large-scale dataset containing `good' and `confusing' images for VPR state-of-the-art. Such a dataset could subsequently help in training an end-to-end neural network for classifying an image as good/bad for map-insertion. We hope that this work draws attention of VPR community towards further research in segregation between confusing and good images.  Thus, moving closer to practical deployment of VPR systems.

{
\small
\bibliographystyle{ieeetr}
\bibliography{root}
}

\begin{IEEEbiography}[{\includegraphics[width=1in,height=1.25in,clip,keepaspectratio]{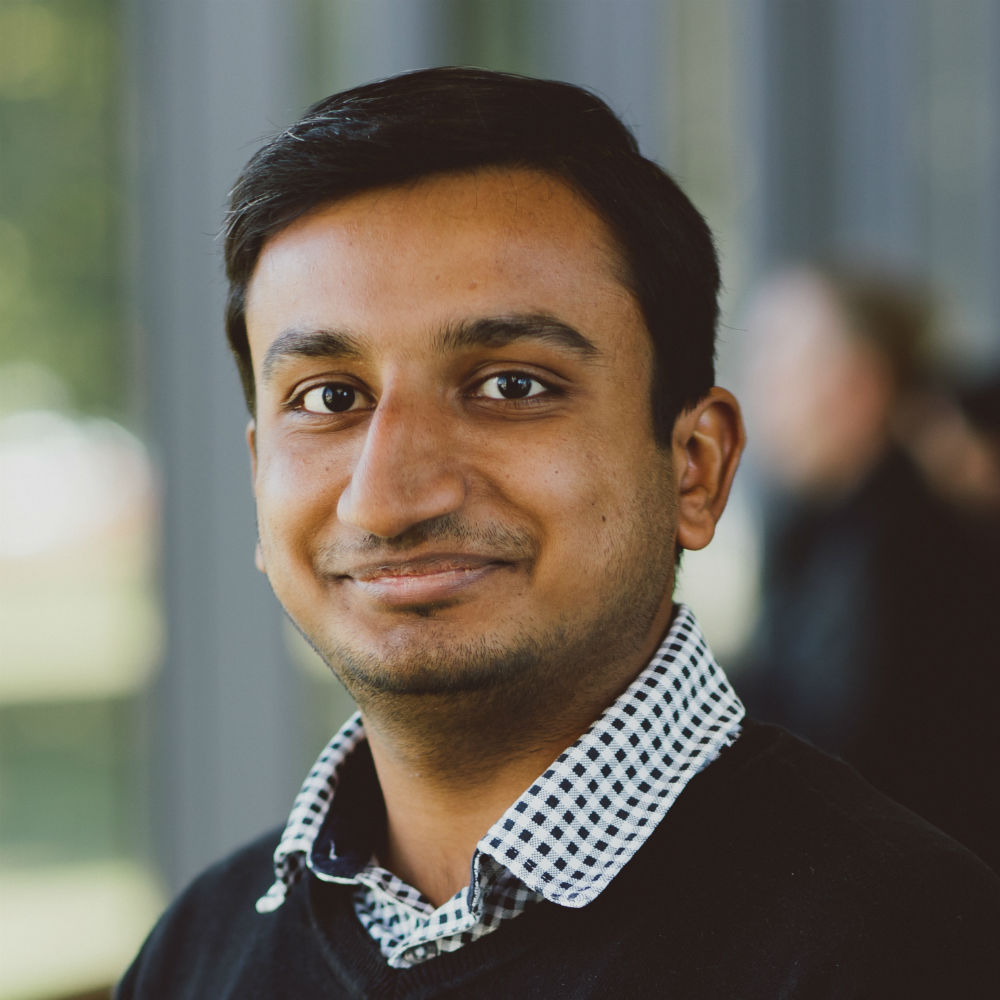}}]{Mubariz Zaffar}
received his BE electrical engineering degree from National University of Sciences and Technology (NUST), Pakistan in 2016. He is the recipient of south-asian helix innovation award, DICE foundation innovation award and NUST high-achiever's award. He is currently pursuing an M.Sc in Computer Science and Electronic Engineering at University of Essex. He is also working as a research officer in the National Centre for Nuclear Robotics (NCNR) project. His research interests include computer vision and deep learning for autonomous robotics, visual place recognition and robot navigation, simultaneous localization and mapping, radiation effects on embedded systems and radiation resilient techniques for nuclear robotics.   
\end{IEEEbiography}

\begin{IEEEbiography}[{\includegraphics[width=1in,height=1.25in,clip,keepaspectratio]{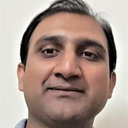}}]{Shoaib Ehsan} (SM'18)
received the B.Sc. degree in electrical engineering from the University of Engineering and Technology, Taxila, Pakistan, in 2003, and the Ph.D. degree in computing and electronic systems with a specialization in computer vision from the University of Essex, Colchester, U.K., in 2012. He has extensive industrial and academic experience in the areas of embedded systems, embedded software design, computer vision, and image processing. His current research interests are in intrusion detection for embedded systems, local feature detection and description techniques, image feature matching, and performance analysis of vision systems. He is a recipient of the University of Essex Post Graduate Research Scholarship and the Overseas Research Student Scholarship. He is a winner of the prestigious Sullivan Doctoral Thesis Prize by the British Machine Vision Association.   
\end{IEEEbiography}

\begin{IEEEbiography}[{\includegraphics[width=1in,height=1.25in,clip,keepaspectratio]{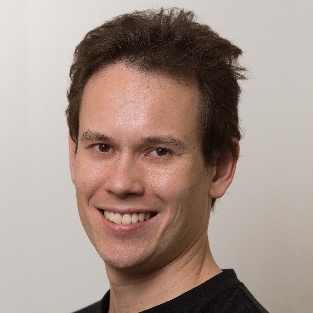}}]{Michael Milford}
(S'06-M'07) received the Ph.D. degree in electrical engineering and the Bachelor's of Mechanical and Space Engineering degree from University of Queensland, Brisbane, Australia. He is currently an Associate Professor and Australian Research Council Future Fellow with Queensland University of Technology (QUT), Brisbane, Australia, and a Chief Investigator for the Australian Centre of Excellence for Robotic Vision. He was a Research Fellow on the Thinking Systems Project with Queensland Brain Institute on the Thinking Systems Project until 2010, when he became a Lecturer with QUT. He conducts interdisciplinary research into navigation across the fields of robotics, neuroscience, and computer vision. Dr. Milford received an inaugural Australian Research Council Discovery Early Career Researcher Award in 2012 and became a Microsoft Research Faculty Fellow in 2013.   
\end{IEEEbiography}

\begin{IEEEbiography}[{\includegraphics[width=1in,height=1.25in,clip,keepaspectratio]{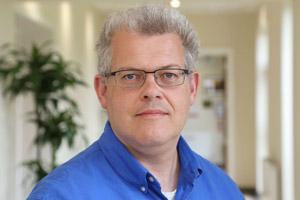}}]{Klaus D. McDonald-Maier}
(SM'12) received the Dipl.-Ing. degree in electrical engineering from the University of Ulm, Germany, the M.S. degree in electrical engineering from the École Supérieure de Chimie Physique Électronique de Lyon, France, in 1995, and the Ph.D. degree in computer science from Friedrich Schiller University, Jena, Germany, in 1999. He worked as a Systems Architect on
reusable microcontroller cores and modules with the Infineon Technologies AG's Cores and Modules Division, Munich, Germany, and a Lecturer in Electronics Engineering with the University of Kent, Canterbury, U.K. In 2005, he joined the University of Essex, Colchester, U.K., where he is currently a Professor with the School of Computer Science and Electronic Engineering. His current research interests include embedded systems and system-on-a-chip design, security, development support and technology, parallel and energy efficient architectures, and the application of soft computing and image processing techniques for real world problems. He is a member of the Verband der Elektrotechnik Elektronik Informationstechnik and the British Computer Society, and a fellow of the Institution of Engineering and Technology.  
\end{IEEEbiography}

\end{document}